\documentclass[runningheads]{llncs}
\usepackage[T1]{fontenc}
\usepackage{graphicx}
\usepackage{amsmath}
\usepackage{booktabs}
\usepackage{multirow}
\usepackage{subcaption}
\usepackage{caption}
\usepackage{wrapfig}
\usepackage{amssymb}
\usepackage{pifont}
\usepackage{colortbl}
\usepackage{hyperref}

\usepackage{color}

\begin{document}
\title{RoLD: Robot Latent Diffusion for Multi-task Policy Modeling}
\titlerunning{RoLD: Robot Latent Diffusion}

\author{
    Wenhui Tan\inst{1}\thanks{This work was performed when Wenhui Tan and Junbo Zhang were visiting Microsoft Research as research interns.} \and
    Bei Liu\inst{2}\and
    Junbo Zhang\inst{3}\and
    Ruihua Song\inst{1}\and
    Jianlong Fu\inst{2}
}
\authorrunning{W. Tan et al.}
\institute{Renmin University of China, Beijing, China \and Microsoft Research, Beijing, China, \and Tsinghua University, Beijing China\\
\email{tanwenhui404@ruc.edu.cn}
}
\maketitle

\vspace{-0.5cm}
\begin{abstract}
Modeling generalized robot control policies poses ongoing challenges for language-guided robot manipulation tasks. Existing methods often struggle to efficiently utilize cross-dataset resources or rely on resource-intensive vision-language models, thus limiting their multi-task performance and practical applications.
In this study, we propose a novel approach that decouples robot action trajectory encoding and control policy generation by leveraging latent action trajectory spaces, enhancing the generalization ability of policy generation on multi-task manipulation tasks. First, we pre-train a task-agnostic auto-encoder to project an action trajectory of several frames accompanied with observations into a latent action trajectory space on large-scale datasets collected with multiple embodiments in various environments. Then we propose learning a diffusion model based on the latent action trajectory space to generate actions of next steps.
Through experiments on two widely used benchmarks, results demonstrate that our proposed method outperforms baselines by 7\%-29\% in terms of average success rate across eight tasks. Our method can consistently benefit from pre-training while baselines cannot. Our method is more than two times faster than our baseline. 

\end{abstract}

\section{Introduction}
\label{sec:intro}
Multi-task robot control policy modeling has a long history within the fields of vision-language and robotics, and attracts more and more interests recently with the great success of large models in the areas of natural language processing~\cite{llama,llmsurvey}, computer vision~\cite{vit,swin}, and vision-language~\cite{clip,lynx}. Training robot control policies requires learning from human-controlled robot demonstrations with parallel data on robot action, observation, and task instructions. Nevertheless, the inherent heterogeneity of robot datasets severely hinder seamless joint training across datasets. The observed heterogeneity stems from a variety of factors, including the distinct models of robots utilized (such as the 7-axis Franka robot and the 6-axis UR robot), the differing styles of robotic operation (e.g., two-fingered robots versus gripper-less robots), and the diverse configurations in their physical appearances and camera frame settings~\cite{rtx}. Therefore, to effectively train generalized robot control policies for multi-tasks, a unified representation of these diverse data is urgently needed.

\begin{figure}
    \centering
    \includegraphics[width=\linewidth]{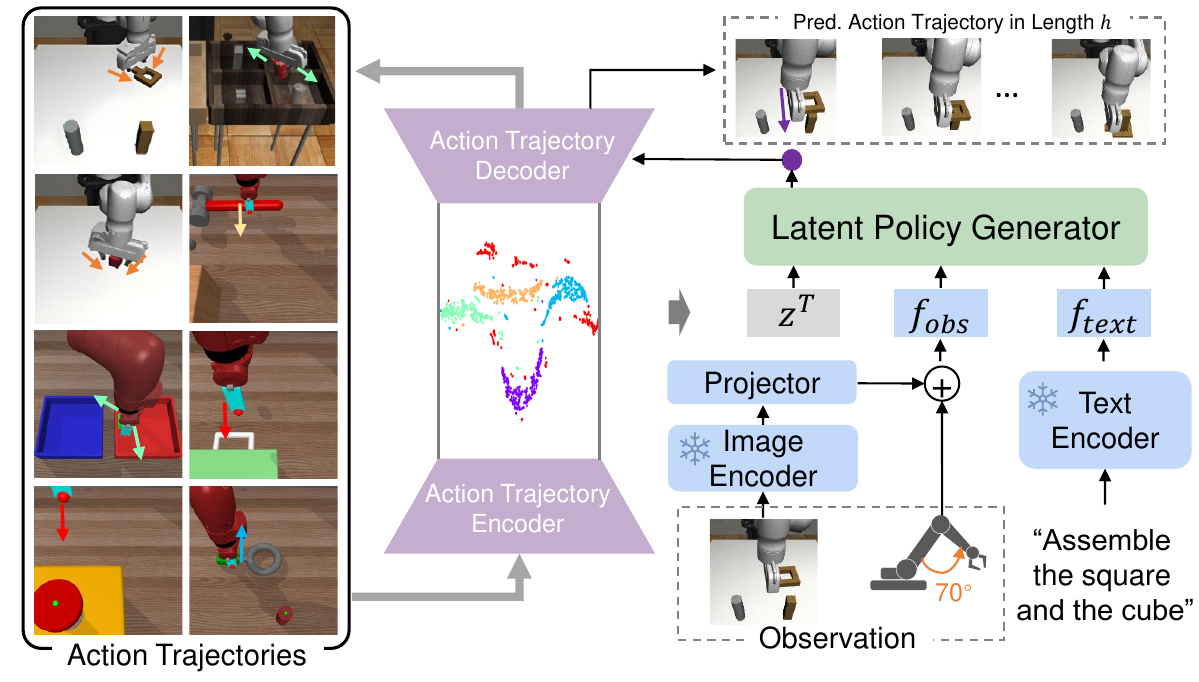}
    \caption{Flowchart of our proposed method.}
    \label{fig:teaser}
\end{figure}
To achieve more stable and effective robot manipulation policy learning process, some previous works have tried using generative models to predict continuous action trajectories~\cite{diffusionpolicy,act}. However, although the works improves single-task performance, they directly generate action trajectories from observations and instructions, without special designs to learn unified representations of action trajectories across tasks, limiting their multi-task effectiveness. Some other works propose to predict discrete action tokens, which are encoded by Vector Quantization method (VQ)~\cite{lisa}, or next control signals in large language models~\cite{rt2}. Despite of the advantage of leveraging internet-scale pre-trained vision-language models for robotics by RT-2~\cite{rt2}, their model size exceeds that of conventional manipulation policy models, limiting their usefulness for real-time inference. Few works investigate how to learn a unified latent representation of action trajectories across tasks and use diffusion based models to generate policies.

In this paper, we propose a novel approach that efficiently encodes action trajectories into a latent space, which is also unified for multi-tasks, and further applies latent diffusion for effective policy generating, namely RoLD (\textbf{Ro}bot \textbf{L}atent \textbf{D}iffusion).
As shown in Fig.~\ref{fig:teaser}, our methodology commences with compressing diverse action trajectory sequences from different embodiments (or robots) and different environments into a compact latent space by an auto-encoder. To achieve high generalization capability, we pre-train the proposed auto-encoder on the large scale Open-X-Embodiment~\cite{rtx} robot datasets collection, with 24 diverse subsets, over 7 various embodiments. 
In the subsequent phase, a latent diffusion process recovers Gaussian noise to the encoded latent action trajectory conditioned on task instructions and observations for effective and efficient multi-task policy modeling. The recovered latent trajectory is then decoded into actions, which are finally executed.
Experimental result show that our proposed method outperforms the best baseline by more than 7\% in terms of average success rate on eight tasks. Further analyses indicate that our model can benefit from pre-training over diverse robotic datasets and learn a unified latent space.  Simultaneously, our model can harness the power of latent diffusion for efficient and precise latent action trajectory predictions. 

In summary, our contributions are as follows:
\begin{itemize}
\item We present unified action trajectory modeling to condense a variety of action sequences into a compact latent space with rich skill-level semantics, enabling effective pre-training across multi-embodiment and multi-task datasets.
\item We propose a latent diffusion policy generator that stably denoises in the latent action trajectory space conditioned by task indicators and observations to enable multi-task manipulation.
\item Through extensive experiments, we demonstrate that our approach surpasses existing baselines by substantial margins of 7\% to 29\% across eight tasks. Results also indicate the consistent effectiveness of pre-training of our method.
\end{itemize}
\section{Related Works}  

\subsection{Robot Manipulation Policy Modeling}
Robot manipulation policy modeling could be rooted in Image-Based Visual Servoing (IBVS)~\cite{visualservo}, which predominantly utilizes manually crafted algorithms. In contemporary research, the focus has shifted towards learning-based policies, now the dominant paradigm. A significant line within this domain encompasses Reinforcement Learning (RL) based approaches~\cite{decision,siegel2020keep}, which integrate policy modeling with RL to notable effect. Despite their advancements, these methods often grapple with challenges such as training inefficiencies and limited generalization with RL. Conversely, current mainstream paradigm to robot policy modeling is learning from human demonstrations~\cite{rt2,generalist,vima,pave} or benefiting from real-world videos~\cite{vc1,r3m,stp}. This paradigm has demonstrated remarkable training efficiency and effectiveness across both simulated and real-world environments. Our work is inline with this approach, aiming to leverage cross-dataset joint training to benefit multitasking.
\vspace{-0.5cm}
\subsection{Generative Models for Policy Modeling}  
As generative models have progressed in computer vision, natural language processing and vision-language~\cite{gan,vae,ldm}, a new wave of generative approaches has emerged in robot learning~\cite{act,unipi}. Notably, Chi et al.~\cite{diffusionpolicy} utilize a diffusion model to engineer a gradient field conducive to manipulation policy modeling, achieving commendable stability and efficacy. However, these methods prioritize single-task proficiency at the expense of multi-task versatility. For instance, Zhao et al.~\cite{act} leverage a variational auto-encoder for trajectory prediction, followed by action chunking to enhance action execution. Our method adopts this paradigm by harnessing latent diffusion for decoupled action encoding and control policy generation, thereby improving multi-task and pre-training capabilities.
\vspace{-0.5cm}
\subsection{Diffusion Models}  
Denoising diffusion probabilistic models (DDPMs)~\cite{ddpm} have recently risen to prominence, rivalling Generative Adversarial Networks (GANs)~\cite{gan} and Variational Autoencoders (VAEs)~\cite{vae}. The DDPMs framework conceptualizes data generation as a diffusion process, thereby offering a distinct perspective on model formulation. Nichol et al.~\cite{ldm} introduce Latent Diffusion Models (LDMs), which amalgamate the merits of VAEs and diffusion models. This innovation involves conducting the diffusion process in a latent space, rather than the traditional pixel space, which not only improves sample quality but also reduces computational requirements.
Our study adopts LDMs to maximize the advantages of independent action trajectory modeling and policy modeling.
\begin{figure}[t]
    \centering
    \includegraphics[width=\textwidth]{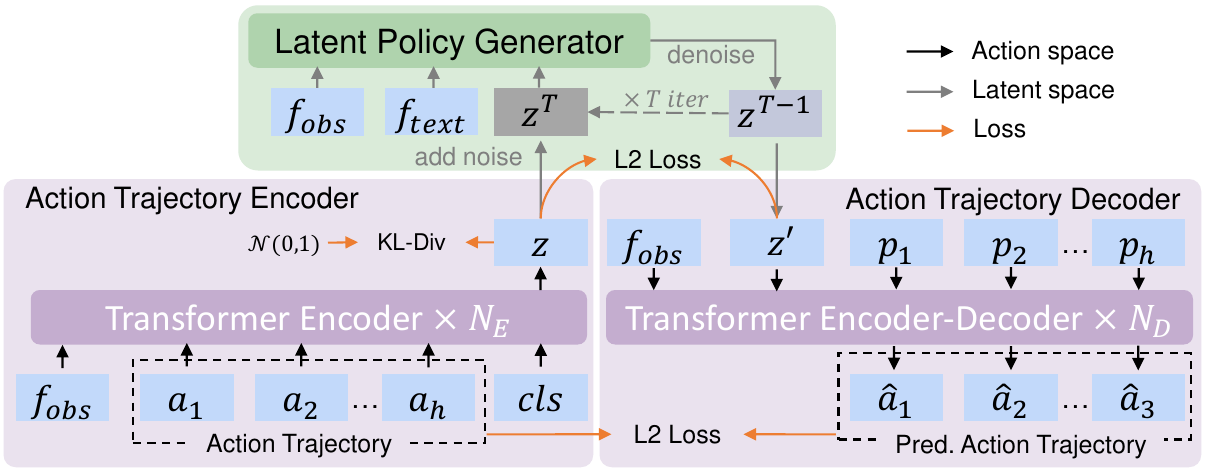}
    \caption{RoLD comprises two core components: 1) a Action Trajectory Auto-encoder (ATA) for unified action trajectory modeling in a condense latent space and 2) a Latent diffusion based Policy Generator (LPG) that iteratively denoises sampled noise to recover target trajectory latent $z$, conditioned by observations and instructions for efficient policy modeling. In this figure, $cls$, $\{a_i\}$ and $\{p_i\}$ denote learnable token, action and position embedding, respectively. $N_E$/$N_D$ represent the number of layers of Action Trajectory Encoder/Decoder. The $f_{obs}$ tokens are identical in this figure.}
    \label{fig:framework}
\end{figure}

\section{Method}
In this section, we describe our proposed new framework RoLD as shown in Fig.~\ref{fig:teaser}. We first delve into the pipeline of preparing large-scale pre-training data in Section~\ref{sec:training}. Then we introduce the two critical components of RoLD: 1) a task-agnostic yet embodiment-aware auto-encoder employed for the unified modeling of the latent action trajectory space, and 2) a task-aware diffusion model, applied for multi-task policy modeling, in Section~\ref{sec:ae} and Section~\ref{sec:ldm}, respectively.

\subsection{Pre-training Dataset Construction}\label{sec:training}  
To harness the wealth of information in large scale diverse robotics datasets for generalized training, we utilize the Open-X-Embodiment dataset~\cite{rtx}. This dataset is critically important for learning a latent space of action sequences and establishing a multi-task policy model. The Open X-Embodiment dataset encompasses data from 70 individual real-world datasets, featuring 22 different robotic embodiments, and is continually expanding with new data. Given the substantial variability in embodiments, instructions, observations, and action spaces among sets, we employ a pre-processing pipeline before pre-training.\\

\textbf{Data Filtering.} We manually check and filter some datasets to ensure the quality of data for pre-training. First, we exclude the datasets irrelevant to our goal, specifically those pertaining to navigational tasks or involving bi-manual robots, as our focus is on tasks involving single-robot manipulation. Second, the datasets with ambiguous action formats or erratic action control signals are omitted to enhance training stability. Finally, pre-training utilizes 24 meticulously subsets from the Open X-Embodiment dataset collection. Data and code could be found in our project \href{https://github.com/AlbertTan404/RoLD}{web-page}.\\

\textbf{Data Normalization.} To achieve more stable training process and facilitate downstream fine-tuning, we design a pipeline to normalize training data. Different from traditional normalization techniques on action data, we first eliminate the outliers that might be caused by unstable physical sensors. Second, we re-scale the remaining action data to the range of [-1, 1]. This normalization is vital for maintaining stability and ensuring compatibility with downstream data, which typically constrains action signals to the same range. Third, for observational data involving multiple camera views, we employ a strategy of random view selection during each training iteration, promoting robust generalization. Fourth, we strictly utilize center cropping without distortion to preserve the fidelity of observation-action pairings. Proprioceptive data (robot states) are not incorporated during pre-training, as their dimensions can be different across datasets and can not be batched. Finally, for downstream tasks, we maintain the original distribution of action data and replicate the image processing approach in pre-training.

\subsection{Action Trajectory Auto-encoder}\label{sec:ae}
As Fig.~\ref{fig:framework} shows, we design a task-agnostic Action Trajectory Auto-encoder (ATA) for unified action trajectory modeling. The introduced latent embedding enables the compression of diverse action sequences of different embodiments under different observations into a concise but semantically rich latent space. 

Specifically, as shown in Fig.~\ref{fig:framework}, we employ a Conditional Variational Auto-Encoder (CVAE)~\cite{cvae} to encode and reconstruct the action trajectory $\mathbf{a}$, using observations features $f_{obs}$ as condition (defined in Eqn.~\ref{eq:obs}). The encoder $\mathcal{E}$ accepts action sequences and observations as inputs, encoding them into an embodiment-aware latent variable, denoted as $z=\mathcal{E}(\mathbf{a},f_{obs})$. The decoder $\mathcal{D}$ conditions on both the latent features $z$ and the image observations to reconstruct the original action sequences, to reconstruct the action trajectory $\hat{\mathbf{a}}=\mathcal{D}(z,f_{obs})$. The Action Trajectory Auto-encoder is trained with two loss terms: 1) an L2 reconstruction loss on action trajectory $\mathbf{a}$ and 2) a KL-penalty towards a standard normal on the learned latent variable:

\begin{equation}
\begin{split}
L_{ATA} = \|\mathbf{a},\hat{\mathbf{a}}\|_2^2 + wKL(q(z|\mathbf{a},f_{obs}) || p(z|f_{obs})),
\end{split}
\end{equation}
where $w$ is the weight applied to KL loss.

During the encoding phase of the Action Trajectory Autoencoder, we process observations including images and proprioceptions from a single frame and the subsequent $h$ frames of action, which we refer to as the horizon of action trajectory. We employ a three-layer transformer encoder to encode the action trajectory and observations in conjunction with a learnable $cls$ token. The encoded feature corresponding to the $cls$ token is reduced to a dimension of $2 \times d_z$, where $d_z$ is the dimension of latent variable $z$. Subsequently the feature is partitioned into two components: $\mu_z$ and $\sigma_z$, standing for the the mean value and standard deviation to parameterize the latent space. This encoder serves to effectively compress the action sequence and encapsulate it within the context of a given observation. In the decoding process (See Fig.~\ref{fig:framework}), a six-layer transformer decoder reconstructs the action trajectory from a sequence of fixed position embeddings $\{p_i\}_{i=1}^h$ conditioned on the concatenated observations and the latent variable $z$ sampled from $\mathcal{N}(\mu_z, \sigma_z)$.

\subsection{Diffusion based Latent Policy Generator}\label{sec:ldm}
By training an Action Trajectory Autoencoder, we obtain a concise and task-agnostic latent action trajectory space. To implement multi-task policy modeling, we further employ a Diffusion based Latent Policy Generator (LPG) to predict the target latent feature conditioned on task instructions and observations, as shown in Fig.~\ref{fig:framework}.
Specifically, our Action Trajectory Auto-encoder (ATA) module's decoder originally accepts sampled noise as input during inference, which mismatches the encoded latent $z$ conditioned with observations during training. Therefore, we aim to design a policy model that can accurately recover the target latent $z$, conditioned on both observations and task instructions, to enable more precise decoding of action sequences.

Diffusion models, a class of generative models, are architecturally conceived to approximate a target data distribution through an iterative denoising process, starting from a Gaussian noise distribution. These models inherently aim to recover an underlying latent distribution by gradually reversing the diffusion process.
Starting from diffusion timestep $t=T$ and a Gaussian noise ${z}^{t}$, DDPM performs $T$ iterations of denoising to recover the original data gradually:
\begin{equation}
z^{t-1}=\alpha(z^{t}-\gamma\epsilon_\theta(z^{t},t)+\mathcal{N}(0,\sigma^2I)),
\end{equation}
where $\epsilon_\theta$ is a network with parameters $\theta$ that predicts the noise and $\mathcal{N}(0,\sigma^2I)$ is the added Gaussian noise. We use the default settings of $\alpha$ and $\gamma$ in DDPM~\cite{ddpm}. In our LPG, the diffusion model is trained to recover the learned latent feature $z$ introduced in Section \ref{sec:ae}. Consequently, the training objective of our model is adjusted to the following form as in latent diffusion model:
\begin{equation}
L_{LDM} = \mathbb{E}_{\mathcal{E}(x),\epsilon\sim\mathcal{N}(0,1), t}\left[||\epsilon-\epsilon_{\theta}(z^t, t)||^2_2\right],
\end{equation}
where $z_t$ is noised from the ATA's encoder $\mathcal{E}(x)$. During inference, the denoised latent feature $z$ is utilized to generate action sequence with ATA's decoder. 

To accurately recover the learned latent trajectory variable $z$ and facilitate multi-task policy learning, observations (including images and robot proprioceptions) and textual task instructions are used as conditions in the denoising process.
Consequently, the objective of LPG is:
\begin{equation}
L_{LPG} = \mathbb{E}_{\mathcal{E}(x),y,\epsilon\sim\mathcal{N}(0,1), t}\left[||\epsilon-\epsilon_{\theta}(z^t, t, \tau_{\theta}(y))||^2_2\right],
\end{equation}
where
\begin{equation}
\tau_{\theta}(y) = concat(f_{obs}, f_{text}, f_t).
\end{equation}

We use a frozen image encoder $\mathcal{F}_v$ and a trainable three-layer MLP to obtain observation features as
\begin{equation}\label{eq:obs}
f_{obs}=MLP(concat(\mathcal{F}_v(img), state)),
\end{equation}
and a frozen language encoder to extract textual task instructions. Diffusion timestep $t$ is encoded into $f_t$ with sinusoidal embedding.

During inference, given current observations and a manipulation task instruction like ``Assemble the square and cube'', we first recover a latent variable $z'$ with our diffusion based latent policy generator, then an action trajectory $\hat{\mathbf{a}}$ is generated with ATA's decoder conditioned on the latent variable and observations.

\begin{figure}[t]
    \centering
    \includegraphics[width=\linewidth]{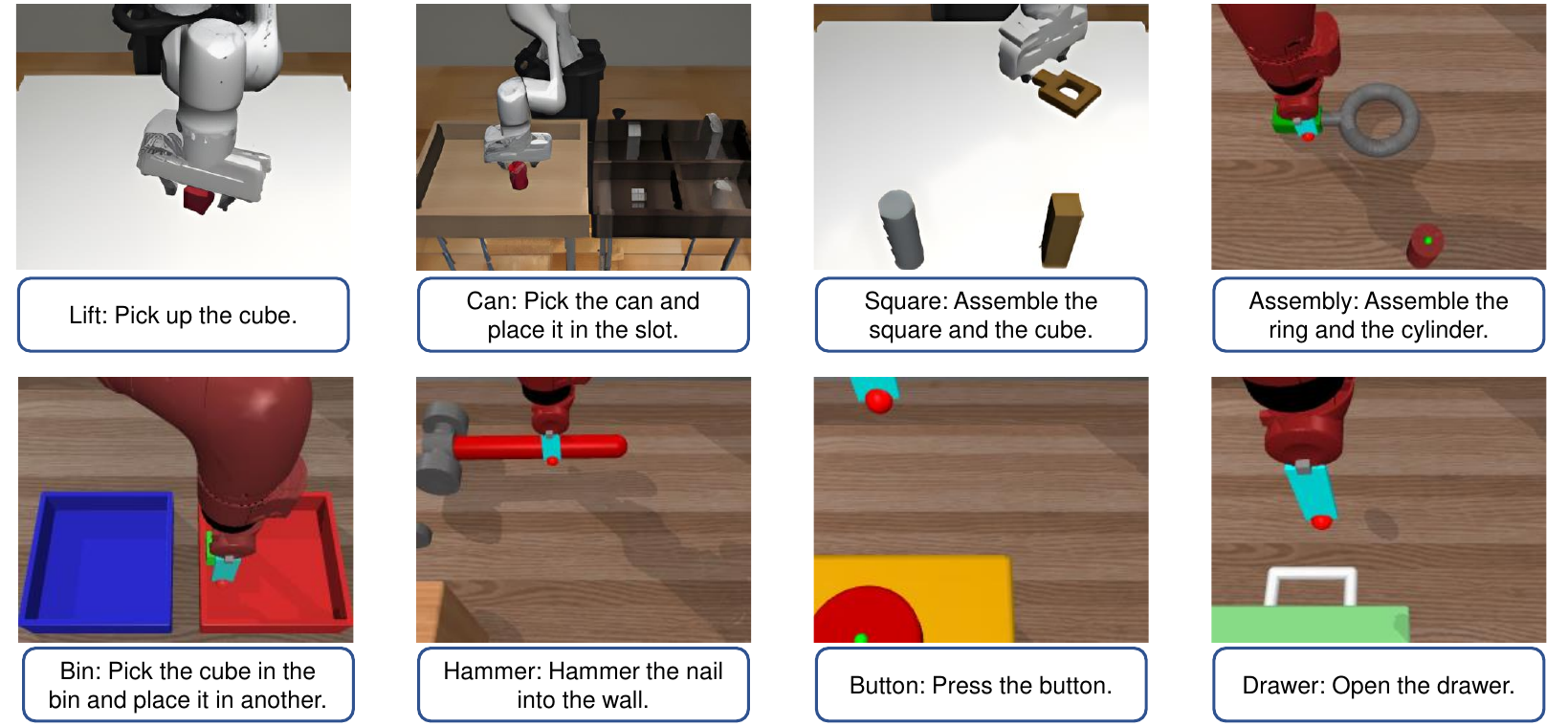}
    \caption{RoLD is able to generalize to diverse tasks conditioned on instructions.}
    \label{fig:datasets_largs}
\end{figure}

\section{Experiments}
We evaluate our proposed method with strong baselines and further analyze contributions of different components, the impact of key parameter, and efficiency.

\textbf{Benchmarks and Evaluation Setup.} Our experiments are based on two widely-used simulation benchmarks: Robomimic and Meta-World. 
1) \textbf{Robomimic}~\cite{robomimic} is a robot-learning benchmark for manipulation tasks including \textit{Lift} (picking), \textit{Can} (picking-and-placing) and \textit{Square} (Assembly). Examples of different manipulation tasks are shown in Fig.~\ref{fig:datasets_largs}. The training data of Robomimic concludes two subsets of better and poorer data quality with suffix -PH and -MH. We separately evaluate the methods trained on the two subsets and average the success rates as the reported results because both have the same comparison conclusions. 2) \textbf{Meta-World}~\cite{metaworld} is an open-source simulation benchmark for robot learning. We adopt the training data and settings of~\cite{vc1} with five manipulation tasks, including \textit{Assembly}, \textit{Button} (button pressing), \textit{Hammer} (hammering), \textit{Bin} (picking-and-placing) and \textit{Drawer} (drawer opening), for multi-task training (See examples in Fig.~\ref{fig:datasets_largs}). During training and evaluation, we do not use any auxiliary information in the simulation environment like the pose of the target object. We set the environment step limit to 400, aligned with~\cite{diffusionpolicy} and average the success rates of 50 trials with different seeds as a final success rate.

\textbf{Training Details.} For visual observation, we employ a frozen R3M-ResNet34 as the backbone of $\mathcal{F}_v$~\cite{r3m,resnet}. For text modality, we encode text tokens with DistilBERT ~\cite{sanh2019distilbert} and average the token features as the final output, ensuring alignment with the R3M framework. KL loss weight $w$ is set to 0.01 in training ATA. We employ a cosine learning rate scheduler with a warm-up period of 1000 steps, peaking at a learning rate of 1e-4. The training dataset is divided into 95\% for training and 5\% for validation. The number of epochs is determined based on the convergence of validation loss. ATA and LPG are trained for 200 epochs and 100 epochs on the downstream tasks respectively, and ten epochs both during pre-training. We use a fixed trajectory unit size of $h=16$ steps considering the average length of training data, i.e., we encode every 16-step action trajectory into a single latent variable $z$.

\textbf{Baselines.} We compare our method RoLD with the following open-sourced state-of-the-art baselines: \textbf{DiffusionPolicy} (DP)~\cite{diffusionpolicy} is a transformer-based diffusion model, which exhibits excellent performance on robot tasks with end-to-end diffusion process. \textbf{ACT}~\cite{act} adopts a VAE as its backbone for trajectory prediction, along with a chunking mechanism for consistent action execution. \textbf{LISA}~\cite{lisa} proposes using the Vector Quantization (VQ) method as bottleneck for low-level action representation learning to boost end-to-end policy modeling. For fair comparison, the training processes of baseline methods are aligned to RoLD and the models are in comparable sizes.

\begin{table}[t]
\centering
\caption{Success rates of RoLD under different settings and baseline methods on Robomimic (average success rates on MH and PH subsets) and Meta-World benchmarks. We use bold and underline to highlight the best  second-best performace. The area in grey shows ablation settings and results of our method.}
\label{tab:sota}
\begin{tabular}{l|ccc|ccccc|c}
\toprule
\multirow{2}{*}{Methods} & \multicolumn{3}{c|}{Robomimic}       & \multicolumn{5}{c|}{Meta-World}                                        & \multirow{2}{*}{Avg.} \\
                         & Lift & Can           & Square        & Assembly      & Button        & Hammer        & Bin           & Drawer &                       \\ \midrule
DiffusionPolicy          & \textbf{1.00} & \underline{0.92}          & 0.17          & 0.96          & 0.52          & \textbf{0.96}          & 0.00          & \textbf{1.00}   & 0.69                  \\
LISA                     & \textbf{1.00} & 0.80          & 0.26          & \textbf{0.98} & 0.76          & \underline{0.94}          &\underline{0.78}          & \textbf{1.00}   & 0.82                  \\
ACT                      & \textbf{1.00} & \underline{0.92}          & \underline{0.34}          & 0.86          & \underline{0.80}          & 0.92          & 0.76          & \textbf{1.00}   & \underline{0.83}                  \\
RoLD (Ours)              & \textbf{1.00} & \textbf{0.99} & \textbf{0.52} & \underline{0.96}          & \textbf{1.00} & 0.86          & \textbf{0.82} & \textbf{1.00}   & \textbf{0.89}         \\ \midrule
\rowcolor[rgb]{0.93,0.93,0.93}- w/o pre-training           & 1.00 & 0.97          & 0.39          & 0.76          & 0.96          & 0.98 & 0.84          & 1.00   & 0.86                  \\
\rowcolor[rgb]{0.93,0.93,0.93}- Non-diffusion LPG            & 0.93 & 0.77          & 0.12          & 0.34          & 0.62          & 0.28          & 0.68          & 0.74   & 0.56 \\
\rowcolor[rgb]{0.93,0.93,0.93}- Task-aware ATA         & 1.00 & 0.90          & 0.16          & 0.74          & 0.86          & 0.90          & 0.76          & 1.00   & 0.79 \\
\rowcolor[rgb]{0.93,0.93,0.93}- Obs-agnostic ATA         & 0.06 & 0.00          & 0.00          & 0.00          & 0.00          & 0.00          & 0.00          & 0.00   & 0.00 \\
 \bottomrule
\end{tabular}
\end{table}

\vspace{-0.3cm}
\subsection{Main Results}
The evaluation results of the eight tasks over two benchmarks are presented in Table~\ref{tab:sota}. The results indicate that our method consistently outperforms the baseline models in terms of multi-task average success rates by a substantial margin from 7\% to 29\%.
Compared to DiffusionPolicy and ACT, our method exhibit  superior performance, which we attribute to our well-structured latent action trajectory space, which facilitates to utilize cross-embodiments pre-training and allows for better generalization.
Our method also surpasses LISA by an average success rate of 7\% with a more effective diffusion process based the latent action trajectories as multi-task policy generator. We also notice that our method dose not exhibit the best performances on all the tasks, such as the task ``Hammer''. We suppose that this is due to significant data distribution difference between pre-training dataset and downstream data, as we observe that our method exhibit much better performance, i.e., 0.98 vs. 0.86 on ``Hammer'' without pre-training.

\begin{figure}
    \centering
    \includegraphics[width=0.6\textwidth]{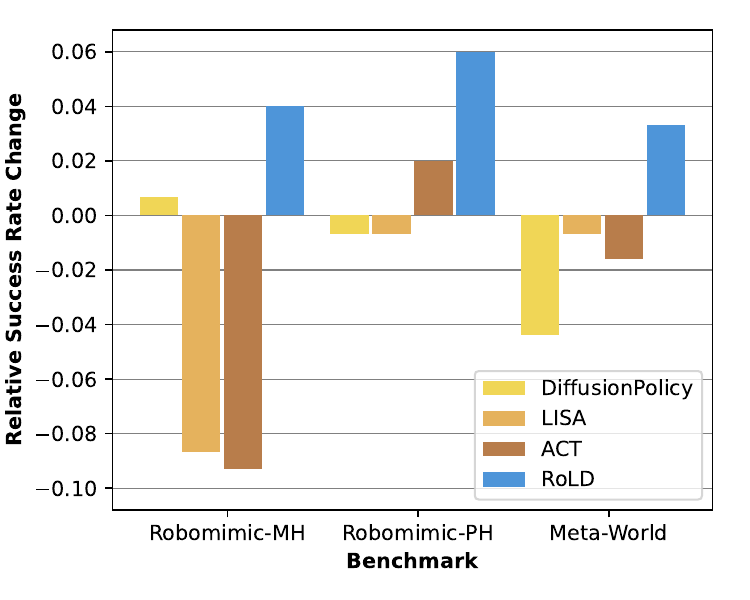}
    \caption{Relative success rate changes of RoLD and baseline methods being pre-trained on our processed Open-X-Embodiments dataset. The methods are evaluated on Robomimic and Meta-World.}
    \label{fig:pt_gain}
\end{figure}

\vspace{-0.3cm}
\subsection{Ablation Study}
We conduct ablation studies to investigate whether our proposed ideas (i.e., pre-training, task-agnostic ATA, and diffusion based LPG) contribute to the best performance and show results in the bottom part of Table~\ref{tab:sota}. The results indicate that all the proposed ideas have dramatic positive contributions. According to the extent of drops, diffusion based LPG contributes the most; the second contributor is task-agnostic ATA; and the third contributor is pre-training. We have the following detailed observations and discussions:

1) When we train a vanilla transformer encoder in the comparable size of LPG to predict the latent variable $z$ instead, significant drops are observed, i.e., 18\% on Robomimic and 38\% on Meta-World in average success rate. This highlights the critical role of the latent diffusion model in accurate multi-task policy modeling.

2) Integrating the autoencoder with task instructions leads to a decline in model performance. This decline is interpreted as a corruption of the pure action trajectory space, underscoring the importance of decoupling action trajectory modeling from policy modeling. 

3) Removing the observation features from ATA results in a collapse of the latent space, thus leading to a deterioration in overall performance, even when LPG is still aware of visual inputs. We attribute this to the intrinsic relationship between trajectories and the robot's observations for modeling action trajectories. For example, an trajectory involving decreasing values along z-axis might signify a ``moving down'' motion if the robot is in contact with a cube or ``assembly'' if the robot is holding a ring.
\vspace{-0.4cm}
\subsection{Evaluation of Pre-training}\label{sec:pt}
Although baselines have no pre-training phase, to make comparison more fair, we fit pre-training into each baseline and use the same pre-training dataset as ours to re-train the baseline models. We then subtract the average success rates of the version with pre-training by those of the version without pre-training on different test sets and present results in Figure~\ref{fig:pt_gain}. The figure shows that only our method RoLD exhibits consistent gains of average 4 points from pre-training. This may be mainly attributed to our latent action trajectory modeling. This approach effectively condenses intricate embodiment-action combinations into a unified representation, thereby facilitating cross-skill and cross-embodiment generalization. In contrast, the baseline methods do not show consistent performance gain with pre-training or even tend to deteriorate. This may be because the baseline methods lack a generalizable and flexible action trajectory modeling module, and thus fail to facilitate multitasking with pre-training. They can perform better when using the training data of a specific downstream rather than leveraging more data from other tasks. 

\begin{figure}[t]
    \centering
    \includegraphics[width=\linewidth]{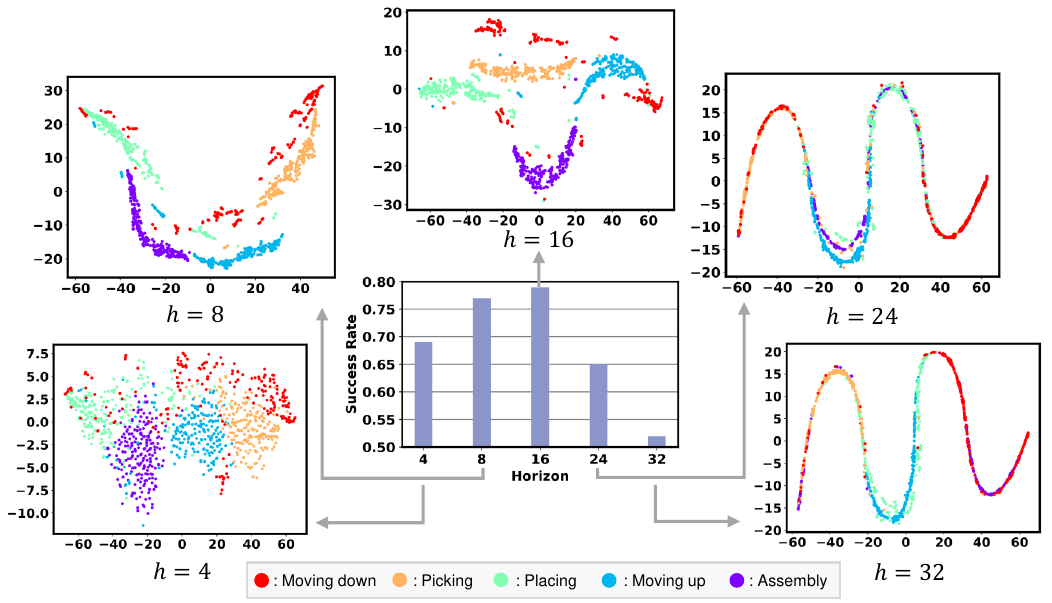}
    \caption{t-SNE visualization of ATA's latent space and success rates on Robomimic with different horizon lengths. We use ATA to encode trajectories from Robomimic dataset into latent variables (data points in this figure), then manually check and label them with different action classes.}
    \label{fig:tsne}
\end{figure}

\vspace{-0.3cm}
\subsection{Impact of Action Trajectory Length $h$}
We conduct further analysis on Robomimic benchmark with how different action horizons, i.e., action trajectory lengths $h$ influence the latent action trajectory space and final success rates. We show the average success rates with $h$ changing and t-SNE visualization of corresponding latent space in Figure~\ref{fig:tsne}. The success rates indicate that as the length of the action trajectory increases, the model's average performance improves, peaking at a length of 16 frames. However, when the length continues to increase, the model's performance dramatically declines.
Accordingly, in the visualized latent spaces when setting action horizon to $h=8$ or $h=16$, we observe that action trajectories tend to cluster better in the latent space, while smaller or larger action horizon exhibit more overlaps. When we manually look into their corresponding action trajectories and label their rough semantics, we find that the dots in the same cluster tend to correspond to the same skills in semantic. We attribute this to the fact that short trajectories typically express weak or no skill-relevant features, making it difficult to construct a unified action space with skill semantics. On the other hand, when the horizon goes larger (24 and 32 frames), more complex movements are included in one dot and they have more overlap with other movements. Thus they are difficult to distinguish from each other, resulting in subpar performance.

\begin{table}[!t]
\caption{Success rates and inference time (seconds per iteration) under various diffusion scheduler settings on RoLD and DiffusionPolicy (DP) on Robomimic.}
\centering
\begin{tabular}{l|c|cccc}
\toprule
Method         & DP   & \multicolumn{4}{c}{RoLD}                  \\
\midrule
Inference Steps & 1000      & 1000      & 250      & 100       & 50            \\
\midrule
Success Rate    & 0.70            & 0.79     & 0.79        & 0.76   & 0.78     \\
Time Cost           &   7.56   &2.26($\times3.3\uparrow$)      & 0.62    &  0.25  &  0.12          \\
\bottomrule
\end{tabular}
\label{tab:speed}
\end{table}

\vspace{-0.3cm}
\subsection{Evaluation of Efficiency}
Given the necessity for real-time inference in robot manipulation tasks, we conduct experiments to evaluate efficiency using different diffusion schedulers and inference timesteps on the Robomimic benchmark. The results are shown in Table~\ref{tab:speed}. We use DDPM as the default diffusion scheduler with 1000 inference steps ($T=1000$) and DDIM~\cite{ddim} for faster inference process ($T=\{250,100,50\}$). The speed is averaged over 100 iterations on a single Tesla A100 GPU. The results in Table~\ref{tab:speed} indicate that our method is as $3.3\times$ faster as DiffusionPolicy while exhibiting superior performance. This improvement can be attributed to our approach of compressing action trajectories into a compact latent space, facilitating a more lightweight diffusion process, which is the most time-consuming part.
Moreover, transitioning to DDIM scheduler with fewer inference steps can further boost inference speed without compromising performance on RoLD.

\vspace{-0.3cm}
\section{Conclusion}
To effectively utilize diverse datasets in robot manipulation tasks and train a more generalized multi-task policy, we proposed a novel framework that pre-trains a latent action trajectory space by using a well designed auto-encoder and generates latent policy by a diffusion model based on the latent space. Experimental results on two widely-used multi-task robot learning benchmarks demonstrate that our method outperforms baselines over 7\% in average success rate and all the proposed ideas have positive contributions. At the same time, our work improves the efficiency by two times over to the diffusion based baseline DiffusionPolicy.  Future work can exploit adaptive horizon length searching for more accurate trajectory latent space modeling, and advanced diffusion processor to accelerate policy modeling.\\
\textbf{Acknowledgement.} This work is supported by the National Natural Science Foundation of China (No. 62276268) and Microsoft Research.

\bibliographystyle{splncs04}
\bibliography{main}

\end{document}